# From Unimodal to Multimodal: Improving sEMG-Based Pattern Recognition via Deep Generative Models

Wentao Wei, Linyan Ren

*Abstract*—*Objective:* **Multimodal hand gesture recognition (HGR) systems can achieve higher recognition accuracy compared to unimodal HGR systems. However, acquiring multimodal gesture recognition data typically requires users to wear additional sensors, thereby increasing hardware costs.** *Methods:* **This paper proposes a novel generative approach to improve Surface Electromyography (sEMG)-based HGR accuracy via virtual Inertial Measurement Unit (IMU) signals. Specifically, we trained a deep generative model based on the intrinsic correlation between forearm sEMG signals and forearm IMU signals to generate virtual forearm IMU signals from the input forearm sEMG signals at first. Subsequently, the sEMG signals and virtual IMU signals were fed into a multimodal Convolutional Neural Network (CNN) model for gesture recognition.** *Results:* **We conducted evaluations on six databases, including five publicly available databases and our collected database comprising 28 subjects performing 38 gestures, containing both sEMG and IMU data. The results show that our proposed approach significantly outperforms the sEMG-based unimodal HGR approach (with increases of 2.15%-13.10%). Moreover, it achieves accuracy levels closely matching those of multimodal HGR when using virtual Acceleration (ACC) signals.** *Conclusion:* **It demonstrates that incorporating virtual IMU signals, generated by deep generative models, can significantly improve the accuracy of sEMG-based HGR.** *Significance:* **The proposed approach represents a successful attempt to bridge the gap between unimodal HGR and multimodal HGR without additional sensor hardware, which can help to promote further development of natural and cost-effective myoelectric interfaces in the biomedical engineering field.**

## I. INTRODUCTION

MYOELECTRIC pattern recognition is a pivotal technology of achieving natural and efficient prosthesis control and human-machine interaction systems of rehabilitation training for biomedical engineering applications. It converts the user's movements into commands for device control and interface manipulation. Surface Electromyography (sEMG) can be indicative of human movement intention, even when collected from the residual limb of amputees [1]. Therefore, sEMG-based pattern recognition systems are widely utilized in many fields such as prosthesis control [2, 3], rehabilitation engineering [4, 5], and non-invasive human-machine interaction [6, 7].

Deep learning (DL) methods have been demonstrated the superior performance in improving the accuracy of sEMG-based pattern recognition [8, 9]. However, these approaches are of low interpretability because the DL-based gesture recognition process is being considered as a "black-box" for myoelectric control, thereby making further improvement of recognition accuracy challenging [10, 11]. The accuracy of sEMG-based pattern recognition systems in recognizing gestures composed of hand and forearm movements remains limited. This can be attributed to 1) the forearm muscles such as the anterior flexors, the posterior extensors, and the medial flexor-pronators mainly control the movement of the wrist and fingers, thus the forearm sEMG signals provide limited information for recognition of complex gestures containing forearm movements [12, 13]; 2) sEMG signals are non-stationary stochastic signal with a high signal-to-noise ratio, such inherent nature can predictably affect the outcomes of gesture recognition [14]. Multimodal HGR has been proposed as an explainable promising approach [15, 16]. Multimodal HGR involves input data streams from at least two modalities that can provide more information for HGR from different dimensions. The multimodal combination of incorporating Inertial Measurement Unit (IMU) signals as an additional information source to sEMG signals is widely recognized [17, 18]. Because the sEMG is an intuitive biosignal capable of capturing subtle distinctions within similar hand gestures whereas the IMU signals is good at capturing hand orientations as well as hand and forearm movements [19, 20]. However, sEMG-IMU-based multimodal HGR systems require additional IMU sensors, which increases the hardware cost [21]. Therefore, how to implement multimodal HGR without additional sensor hardware is of great significance for cost-effective myoelectric interfaces in biomedical engineering applications.

Deep generative models such as Generative Adversarial Networks (GAN) are able to generate high-dimensional and

This work has been submitted to the IEEE for possible publication. Copyright may be transferred without notice, after which this version may no longer be accessible. This work was supported in part by the National Natural Science Foundation of China under Grant 62002171, in part by the Natural Science Foundation of Jiangsu Province under Grant BK20200464, and in part by the Postgraduate Research & Practice Innovation Program of Jiangsu Province under Grant KYCX23_0430.

The authors are with the School of Design Arts and Media, Nanjing University of Science and Technology, Nanjing 210094, China (correspondence e-mail: weiwentao@njust.edu.cn).

complex data by learning the intrinsic correlation and latent distribution of the training data [22, 23]. Many of the currently available sEMG benchmark databases, such as the subdatabases of NinaPro, provide synchronously captured forearm IMU signals data [24-28]. By exploring the intrinsic correlation between sEMG signals and IMU signals of performed gestures and training a deep generative model, it becomes feasible to generate corresponding virtual forearm IMU signals from the input forearm sEMG signals, thus achieving multimodal HGR systems that use sEMG signals and generated virtual IMU signals as inputs.

The major contributions of our work are as follows:
- We propose a novel generative multimodal HGR approach based on generative learning, which can bridge the gap between the unimodal HGR and multimodal HGR without additional sensor hardware.
- We train a deep generative model based on the intrinsic correlation between forearm sEMG signals and IMU signals, and then generate virtual forearm IMU signals by feeding sEMG signals into the trained deep generative model.
- We created a database by measuring muscle activity for various gestures under two distinct forearm states. Experimental results on five publicly available databases and our collected database demonstrate that incorporating virtual IMU signals can significantly improve the accuracy of sEMG-based HGR.

The rest of this paper is organized as follows. A review of the related work is presented in Section II; section III outlines the formulation of the generative multimodal learning problems; section IV introduces the details of the proposed approach; section V demonstrates the experimental databases, experimental configuration, and experimental results; section VI concludes the paper and discusses future work.

## II. RELATED WORKS

In recent years, significant advancements have been made in the performance of sEMG-based unimodal HGR systems through various approaches such as manually constructing more effective feature sets [29, 30] and training advanced deep learning models with higher performance [31, 32]. As sEMG is a high-noise, non-stationary random signal, previous research typically extracted hand-crafted features from the sEMG signals for gesture recognition rather than directly using the raw sEMG signals. Feature extraction is employed to highlight positive information in the sEMG signal whereas mitigating the impact of noise and negative sEMG signals [30]. The time-domain (TD), frequency-domain (FD), or time-frequency-domain (TFD) features are manually extracted from the raw sEMG signal [16, 33]. However, the accuracy of HGR is directly affected by the validity of the selected features.

With the increase of sEMG databases and the development of deep neural network technology, an increasing number of researchers are focusing on deep learning-based sEMG HGR approaches [34, 35]. Compared to traditional shallow learning models, the main advantage of deep learning is its strong feature learning ability, which enables automatic feature learning and the construction of end-to-end sEMG HGR systems without relying on manual feature extraction and selection processes [30]. Deep learning models, such as Convolutional Neural Network (CNN), have demonstrated superior performance in the sEMG-based HGR system. For instance, Shen et al [36] designed a CNN-based model for sEMG signal gesture classification, which outperformed Linear Discriminant Analysis (LDA), Support Vector Machine (SVM), and Random Forests on the Ninapro DB5 database by 5.02%, 6.61%, and 5.47% in accuracy, respectively. Inspired by the success of CNN in image classification, feature learning approaches based on CNN have been extensively explored in HD-sEMG-based HGR and shown promising results, that is, surpassing 95.00% accuracy [38, 39]. However, experimental results on large-scale multichannel databases show that the performance of these unimodal (sEMG signals only) HGR systems based on the deep learning models is still limited in recognizing gestures composed of hand and forearm movements. For example, Geng *et al.* [37] introduced an end-to-end CNN model (i.e., GengNet) which achieved 99.60% gesture recognition accuracy using HD-sEMG but reached only 77.80% accuracy with sparse multichannel sEMG. Notably, a pioneering novel Convolutional Vision Transformer (CviT) with stacking ensemble learning proposed by Shen *et al.* [38] achieved gesture recognition accuracies of 80.02% (with a 200 milliseconds window length) on the NinaPro DB2 and 76.83% and 73.23% on Exercise A and Exercise B of NinaPro DB5, respectively. Additionally, many existing deep learning models are regarded as black-box models, making it difficult for us to ascertain the specific mechanisms responsible for improving gesture recognition accuracy [16, 39, 40]. This making further improvement of sEMG-based unimodal HGR accuracy challenging.

The advancement of hardware technology has enabled an increasing number of human-machine interaction devices to support multimodal perception. Newly emerging myoelectric interface products such as the CTRL-labs armband [41] and eCon armband [42] have integrated sEMG electrodes and IMU to support simultaneous sensing of sEMG signals and motion sensing signals. Previous researches have indicated that multimodal HGR can achieve higher recognition accuracy in recognizing gestures composed of hand and forearm movements. Take the NinaPro database as an example, which include basic finger movements as well as wrist, grasp and functional movements performed by the hand and forearm. Duan *et al.* [17] proposed a novel and practical sEMG-Acceleration (ACC)-based hybrid fusion (HyFusion) model, and achieved high accuracy of 94.73%, 89.60%, and 96.44% on the NinaPro DB2, DB3, and DB7 databases, respectively. Shen *et al.* [43] proposed a flexible and modular approach utilizing sEMG and ACC signals, and recognition accuracy on 49 hand gestures (NinaPro DB2) is 94.24%, significantly surpassing the accuracy of the approach only using sEMG signals.

Typically, multimodal HGR requires additional sensor hardware for acquiring another modal signal, which increases hardware cost. Deep generative models are designed to generate new data samples that are indistinguishable from the real data [44]. As a powerful tool for generating complex data, deep generative models such as GAN [23] and its variants [45-47], Variational Autoencoder (VAE) [48], Diffusion Model [49] are widely used in the fields of image synthesis [50-52], natural language processing[53] and audio generation [54, 55]. Based on the properties of the generative model, we try to train a GAN-based generative model to generate virtual forearm IMU signals rather than directly collecting forearm IMU signals using additional sensors in this work.

Previous research has already noticed the potential of generative models in gesture recognition. To improve the accuracy of sEMG-based HGR. For instance, Hu et al [21] proposed a hybrid approach that combines real sEMG signals with corresponding virtual hand poses, resulting in an accuracy improvement of 5.2% on the sparse multichannel sEMG databases. However, the hand poses mainly reflect the subtle motion of the finger and do not provide information about the forearm posture, thus resulting in limited accuracy improvement in recognizing gestures composed of hand and forearm movements [56]. IMU signals present a viable option as they can offer information hand orientations and hand and forearm movements [23, 24]. Moreover, the hand poses include a relatively large number of channels (22-channel), which increases the complexity of the model [24, 25]. It is worth noting that IMU signals contain different types of signals with different numbers of channels, such as ACC signals (36-channel or 3-channel in this work) and Euler angle signals (3-channel in this work). Researchers can select the appropriate IMU signals based on their specific needs and device configurations. Motivated by this, we present a novel deep generative multimodal HGR approach in this paper. Our aim is to improve the accuracy of sEMG-based HGR with generated virtual IMU signals, which can implement multimodal HGR without additional sensor hardware.

## III. PROBLEM STATEMENT

The problem of sEMG-IMU-based multimodal gesture recognition can be formulate as:

$$z = H(m_{sEMG}, m_{imu}; \theta) \quad (1)$$

where $z$ represents classification results, $H$ is the multimodal gesture recognition model used to process the sEMG signals $m_{sEMG}$ and IMU signals $m_{imu}$, $\theta$ is the parameter of $H$. To achieve multimodal HGR without additional sensor hardware, virtual forearm IMU signals $m_{v\_imu}$ instead of real forearm IMU signals were used in this work, see (2).

$$z = H(m_{sEMG}, m_{v\_imu}; \theta) \quad (2)$$

The problem of sEMG-based HGR by generative multimodal approach can be divided into two parts. Firstly, we train a deep generative model to generate virtual forearm IMU signals based on the corresponding input forearm sEMG signals. Then, we train a multimodal CNN for sEMG-virtual IMU-based multimodal pattern recognition.

### A. Deep generative model for multimodal signal generation

As mentioned above, the deep generative model in this paper is built based on GAN. GAN-based deep generative model consists of a generator $G$ and a discriminator $D$, as shown in Fig. 1. Given a segment of an forearm sEMG signals of $k$ frames (denoted by $m_{sEMG} = [s_1, s_2, \cdots, s_k | s_i \in R^{C_1}]$, where $C_1$ is the number of sEMG signal channel), the generator $G$ takes the forearm sEMG signals $m_{sEMG}$ as input and outputs the virtual forearm IMU signals (denoted by $m_{v\_imu} = [a'_1, a'_2, \cdots, a'_k | a'_i \in R^{C_2}]$, where $C_2$ is the number of IMU signal channel), see (3).

$$m_{v\_imu} = G(m_{sEMG}; \theta_g) \quad (3)$$

where $m_{v\_imu}$ is the virtual forearm IMU signals, $G$ is the generative model used to generate virtual forearm IMU signals $m_{v\_imu}$ from the input forearm sEMG signals $m_{sEMG}$, $\theta_g$ is the parameter of $G$.

The discriminator $D$ is a binary classifier that takes real IMU signals $m_{imu}$ and generated virtual IMU signals $m_{v\_imu}$ as inputs $m$, and outputs the probability distribution $D(m)$ of samples coming from real rather than generated ones. The generator $G$ and discriminator $D$ alternate in an adversarial fashion during training. Specifically, the generator $G$ aims to produce more realistic data, whereas the discriminator $D$ attempts to better distinguish between real and generated data. The objective function in training is shown in (4), and the optimization goal is to achieve Nash equilibrium [57], at which point the generator model $G$ is believed to have captured the distribution of real samples [58].

$$\min_G \max_D V(D, G) = \\ E_{x \sim P_t(m_{imu})}[\log D(x)] + E_{y \sim P_t(m_{sEMG})}[\log(1 - D(G(y)))] \quad (4)$$

where $D(x)$ denotes the probability that the discriminator $D$ judges the input data $x$ as a real sample, $x$ is the data from the real IMU signal distribution $P_t(m_{imu})$; $D(G(y))$ denotes the probability that the discriminator $D$ judges that the generated sample $G(y)$ as a real sample, $y$ is the data from the sEMG signal distribution $P_t(m_{sEMG})$.

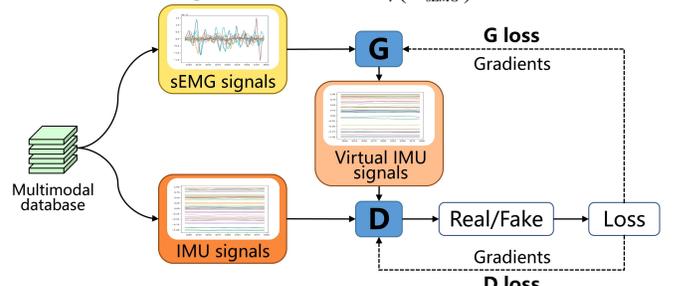

Fig. 1. Illustration of proposed GAN-based deep generative model.

### B. Multimodal CNN for multimodal pattern recognition

Multimodal HGR necessitates the concurrent input of multiple modal signals, for instance, sEMG signals and virtual IMU signals. Thereby, a multimodal CNN model need requires multiple branches to accommodate these signals, and then output the classification result through the integrated fusion of these branches. This multimodal HGR process can be formulated as follows:

$$z = H_f(H_s(m_{sEMG};\theta_s), H_i(m_{v\_imu};\theta_i);\theta_f) \quad (5)$$

where $z$ denotes classification results presented as softmax scores from the multimodal CNN, $H_s$ is the CNN branch used to process the forearm sEMG signals $m_{sEMG}$, $\theta_s$ is the parameter of $H_s$, $H_i$ is the CNN branch used to process the virtual forearm IMU signals $m_{v\_imu}$, $\theta_i$ is the parameter of $H_i$, $H_f$ is the fusion model that fuses multiple CNN branches, $\theta_f$ is the parameter of $H_f$.

## IV. PROPOSED APPROACH

The proposed generative multimodal HGR approach is implemented using a generative multimodal model, which includes the GAN-based deep generative model and multimodal CNN model, as shown in Fig. 2.

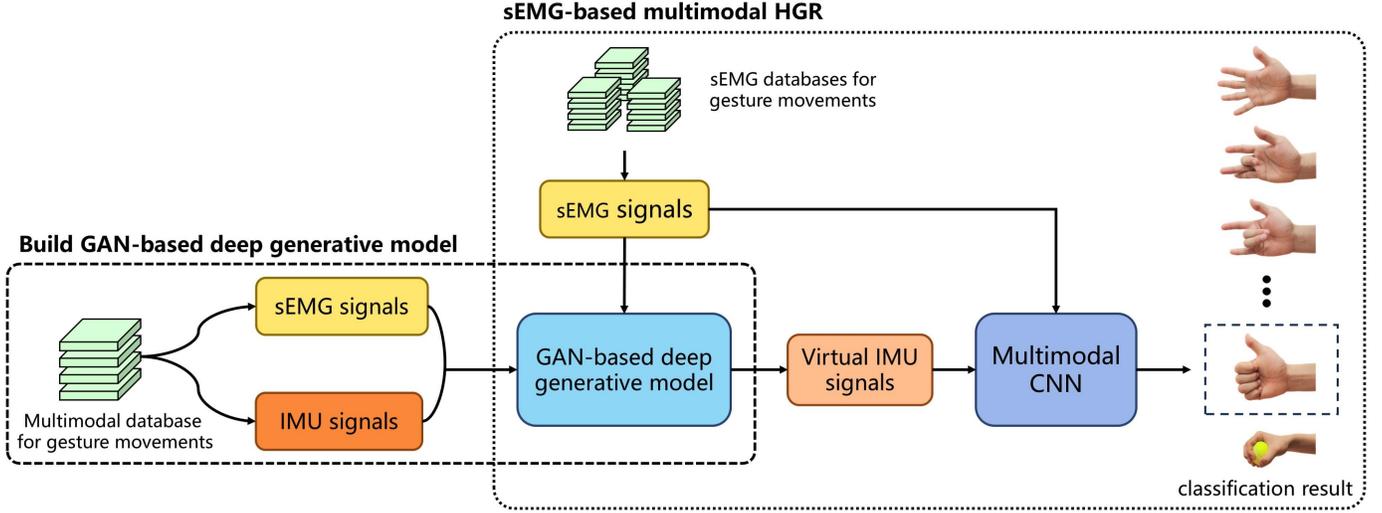

Fig. 2. Illustration of proposed generative approach.

### A. GAN-based deep generative model

The neural network structure of proposed GAN-based deep generative model is shown in Fig. 3. We utilized Deep Convolutional Generative Adversarial Networks (DCGAN) to generate virtual forearm IMU signals from the input forearm sEMG signals in this work. DCGAN is a powerful deep generative model proposed by Radford *et al.* in 2015 [46], which is a variant of Generative Adversarial Networks (GAN). Compared with the GAN, the DCGAN model improves the quality of generated samples and the convergence speed by replacing the original nonlinear maps with convolutional layers and transposed convolution layers [59].

As shown in Fig. 3, the generator $G$ consists of three transposed 2D convolutional layers and one fully connected (FC) layer. Three transposed 2D convolutional layers consist of 32, 16 and 1 3×3 filters with a stride of (1, 2), respectively.

The output sEMG signal features are then flattened and fed into a FC layer, with a number of neurons equivalent to the desired number of virtual IMU signals to be generated. Batch Normalization (BN) [60] and rectified linear unit (ReLU) nonlinearity function [61] are adopted to each transposed convolutional layer, and tanh activation function is adopted to the FC layer.

The discriminator $D$ consists of one 2D convolutional layers and one FC layer. The convolutional layers consist of 16 3×3 filters with a stride of 3. The output sEMG signal features are then flattened and fed into a FC layer, which consist of 1 hidden unit. The BN and LeakyReLU activation function [62] are applied to the convolutional layer, and dropout [63] is applied to prevent overfitting. The Sigmoid is used as the activation function on the FC layer, i.e., the output layer.

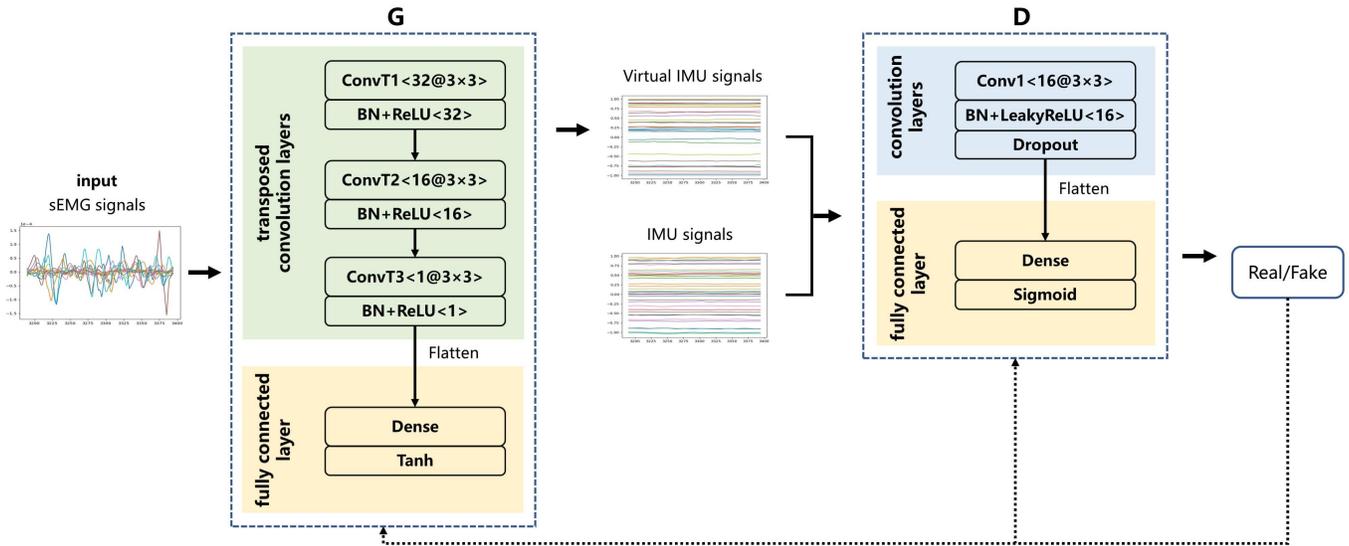

Fig. 3. Illustration of neural network structure of proposed GAN-based deep generative model.

## B. Multimodal CNN

The proposed multimodal CNN presented in this work is a dual-stream CNN that used sEMG and virtual IMU as its multimodal signal inputs. As illustrated in Fig. 4, each stream of the CNN comprises two convolutional layers, two locally connected (LC) layer, and one fully connected (FC) layer. Each convolutional layer is composed of 64 3×3 filters with a stride of 1 and LC layer is composed of 64 1×1 filters with a stride of 1. Subsequently, the outputs from each stream are flattened and separately input into an FC layer with 512 hidden units. Studies have shown that the LC layer imposes an inductive bias by constraining certain features to only appear in specific regions of the input space [64], with filters do not share weights [65]. Compared to the convolution layer, the LC layer has better learning capabilities for local features in images. To accelerate the training process and enhance convergence speed, the BN and ReLU nonlinearity function are adopted to the convolutional layers and LC layers [60, 61]. Additionally, the BN is added before the first convolutional layer in each CNN stream, and dropout is applied to the last FC layer to prevent overfitting.

The outputs of these multiple CNNs are then fused to obtain the final classification results. The fusion module comprises two FC layers. Initially, the output feature vectors from each CNN stream are concatenated and input into an FC layer with 512 hidden units. The ReLU nonlinearity functions is added preceding the FC layer, whereas BN and ReLU nonlinearity functions are applied following the FC layer. Subsequently, a G-way FC layer with softmax activation is utilized to obtain the classification results, where G represents the number of gesture categories.

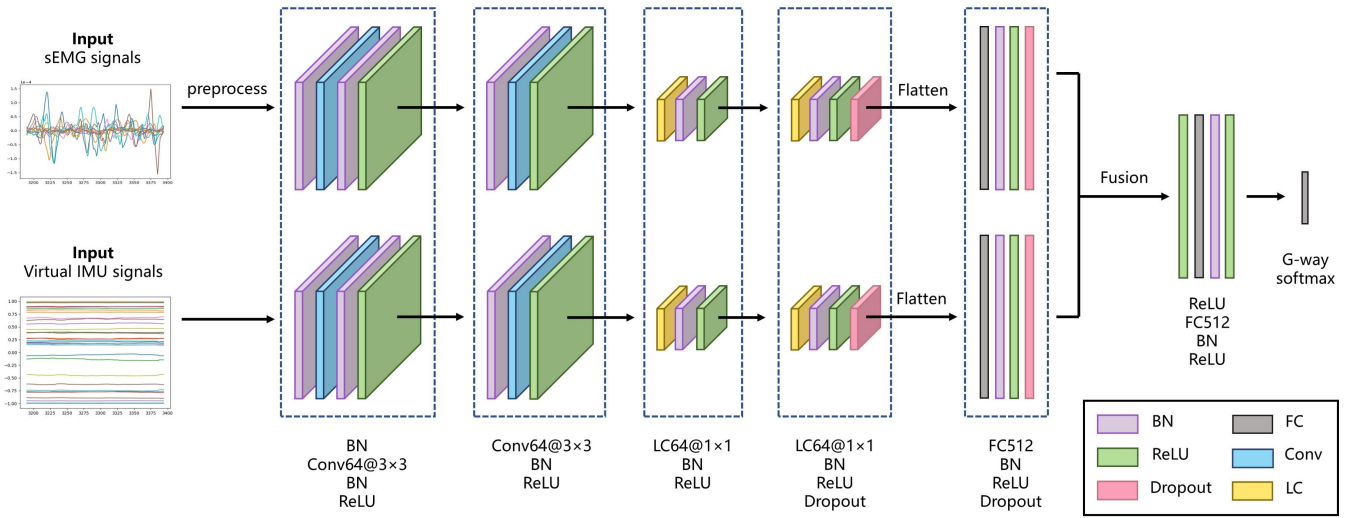

Fig. 4. Illustration of neural network structure of proposed multimodal CNN.

## V. EXPERIMENTS AND RESULTS

We first carried out dataset preparation, signal processing, and experimental setup in this section. Then, to validate the effectiveness of our proposed approach in enhancing gesture recognition performance by incorporating virtual IMU signals, the HGR performance of the proposed generative multimodal HGR approach is compared with the unimodal HGR approach that uses only sEMG as input. Finally, we conduct a performance comparison with state-of-the-art sEMG-based approaches. Six databases utilized for model training and evaluation in this work include our collected database and five publicly available databases (four subdatabases of the NinaPro database and SIEM database). The database specifications are outlined in Table I.

TABLE I
SPECIFICATIONS OF THE DATABASES UTILIZED IN THIS WORK

| Name | Intact subjects | Amputated subjects | Number of subjects to be classified | Number of gestures to be classified | Number of sEMG channels | Number of IMU channels ACC | Number of IMU channels Euler | Number of trials | Number of trials to be classified | Sampling rate |
|---|---|---|---|---|---|---|---|---|---|---|
| FEMG-VPF Database | 28 | 0 | 28 | 38 | 8 | —— | 3 | 6 | 4 | 2040Hz |
| NinaPro DB2 [24] | 40 | 0 | 40 | 50 | 12 | 36 | —— | 6 | 6 | 2000Hz |
| NinaPro DB3 [24] | 0 | 11 | 6 | 50 | 12 | 36 | —— | 6 | 6 | 2000Hz |
| NinaPro DB5 [25] | 10 | 0 | 10 | 53 | 16 | 3 | —— | 6 | 6 | 200Hz |
| NinaPro DB7 [26] | 20 | 2 | 20 | 41 | 12 | 36 | —— | 6 | 6 | 2000Hz |
| SIEM Database [66] | 20 | 0 | 20 | 12 | 8 | —— | 3 | 18 | 6 | 2040Hz |

### A. Data acquisition

To validate the performance of our proposed approach in recognizing gestures composed of hand and forearm movements, we created a database, named "ForeArmEMG-VariedPosForce", denoted as "FEMG-VPF" database. The FEMG-VPF database contains the 18 hand gestures and one rest gesture under two distinct forearm states. Our FEMG-VPF database is now publicly available for research purposes [67], and interested researchers can access the database through the Figshare data repository at https://doi.org/10.6084/m9.figshare.23850366.v1.

*1) Subjects:* The FEMG-VPF database contains data obtained from 28 intact subjects (12 females and 16 males, age: 23.00 ± 2.83 years, 26 right-handed and 2 left-handed). More details about the subjects can be found in [67]. All subjects exhibited no documented skeletal or neuromuscular disorders with no previous myoelectric control experience. The subjects were recruited from the Nanjing University of Science and Technology in China. A brief description of the data acquisition procedure was provided to potential individuals, and those who demonstrated interest was provided full information including methods, purposes, and protocols of data acquisition and subsequently arranged for a session dedicated to data acquisition. All subjects provided written informed consent form before proceeding with the experiments. All protocols adhered to the principles of the Declaration of Helsinki and conformed to applicable regulations within China.

*2) Data acquisition setup:* The sEMG signals and Euler angles in the IMU signals of the forearm muscles were recorded by a commercially available Myo armband (Talmic Labs, Canada) with eight bipolar dry-electrode channels [66]. This Myo armband operated at a sampling rate of 200Hz with an A/D resolution of 8 bits and employs highly accurate zero-drift micropower operational amplifiers (TSZ124, STMicroelectronics) boasting a CMRR of 115 dB and a GBP of 400 kHz. Notably, the FEMG-VPF database also includes forearm HD-sEMG signals collected with a portable sEMG amplifier (Sessantaquattro, OT Bioelectronica, Italy) with 64-channel HD-sEMG electrode grid (OT Bioelectronica, Italy) placed on the forearm surface, sampled at a rate of 2040Hz. The HD-sEMG and sparse multichannel sEMG were recorded synchronously, hence, the sEMG signals from the Myo armband were recorded at a synchronous rate 2040Hz. The data collection process both Myo armband and HD-sEMG is completely non-invasive, thus it does not cause any discomfort or harm to the subjects [68, 69]. To ensure a high correlation between the sEMG and Euler angles used in this study, we exclusively used the sparse multichannel sEMG signal and corresponding Euler angles in the IMU signals gathered by the Myo armband. Despite its comparatively lower sampling rate and number of channels, the Myo armband exhibits a comparable HGR accuracy to other devices, as reported by Pizzolato et al. [25]. In addition, several previous studies have demonstrated the effectiveness of the MYO armband in gesture recognition applications [70, 71].

Each subject was seated on a chair with back upright and shoulders relaxed, positioned in front of a 40-inch LCD monitor. Supervised by a self-developed Qt Program, the laptop facilitated data acquisition.

The two acquisition devices functioned through separate threads, and batches of data from both devices were synchronized and fetched from the computer's random access memory (RAM). Raw data acquisition occurred on a 64-bit Microsoft Windows 10 PC equipped with an Intel i7 1.73-GHz processor and 8-GB RAM.

*3) Data acquisition procedure:* The FEMG-VPF database encompasses 38 gestures composed of hand and forearm movements. The hand gestures were performed under two distinct arm postures: 1) the upper arm naturally drooping with the forearm parallel to the ground, and 2) the upper arm naturally drooping with the forearm at a 45-degree angle to the ground. Each arm posture incorporates one resting gesture (i.e., 0 rest gesture), 15 finger movements (i.e., 1-15) and 3 grasping gestures (i.e., 16-18). Detailed descriptions of the hand gestures and the arm postures are shown in Fig. 5(a) and Fig. 5(b), respectively. The finger movements encompassed in the database include single-finger movement, double-finger movement, triple-finger movement, and full-hand movements. The grasping gestures incorporate actions grasping cylindrical objects (i.e., mineral water bottles), planar objects (i.e., books), and spherical objects (i.e., tennis balls), respectively. Each gesture was performed 6 repeated trials, namely four high-force trials and two low-force trials. Give that the effect of incorporating virtual IMU signals on the gesture recognition accuracy is considered in this work, we utilized the data solely from four high-force trials to mitigating the impact of force variation on the outcomes. The subjects were instructed to first complete all gestures in the 0° arm posture before proceeding with those in the 45° arm posture, and always start and finish trials consistently from the resting position (maintaining fixed arm posture and relaxed fingers). The acquisition time of each trial is 6 seconds. Rest for the first 1 seconds, then start gesturing. As shown in Fig. 6, a 3-second action gesture and a 0.5-second rest gesture were selected to represent stable static gestures in this work. Subsequently, the rest gesture signals from the 18 action gestures in each arm posture were extracted and spliced, totaling 9 seconds. Each subject had a break after completing every six gestures, and they were also permitted to rest at any other point during the experiment [66].

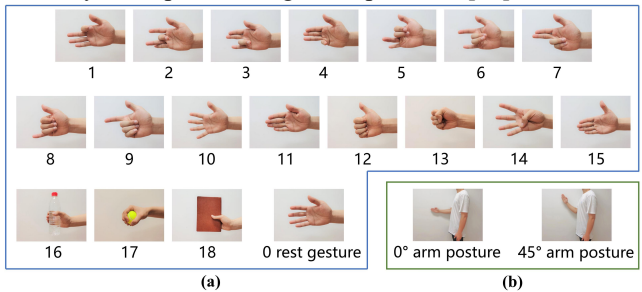

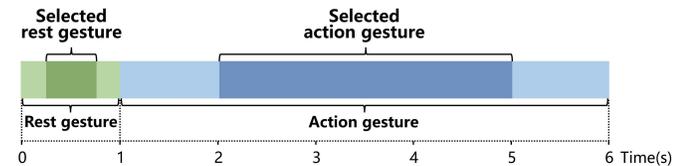

Fig. 5. The specification of the FEMG-VPF database. (a) Eighteen hand gestures and the rest gesture. (b) Two arm postures.

Fig. 6. The movement trajectory and selected gesture data of each trial in this work.

### B. Publicly available databases

*1) NinaPro database:* The NinaPro database is a publicly available large-scale gesture movement benchmark database that include various categories of hand gesture such as basic finger movements, basic wrist movements, and grasp and functional movements. In our work, the proposed approach was evaluated on four subdatabases from the NinaPro database (referred to as NinaPro DB2, NinaPro DB3, NinaPro DB5, and NinaPro DB7) [24-26]. As shown in Table I, all these databases comprise sEMG signals and the corresponding ACC signals sensed by IMU. Referring to the experimental setup of Wei *et al.* [16], we excluded subject data from the aforementioned databases that lacked complete channel and action information.

*2) SIEM database:* SIEM database is a hybrid sEMG signal database presented by Hu et al [66], comprising 20 able-bodied subjects performed 12 finger movements across two paces and three arm postures. The database contains HD-sEMG signals captured through a 64-channel grid positioned on the back of hand, sparse 8-channel sEMG signals and 3-channel IMU signals (Euler angles) recorded by a Myo armband around the forearm, all at a sampling rate of 2040Hz. Consistent with the FEMG-VPF dataset, the sparse 8-channel sEMG signal and corresponding 3-channel Euler angles gathered by the Myo armband used in this work. During three separate experimental sessions, each participant was tasked with conducting a series of three fast and three slow trials consecutively. These trials were performed under various arm postures. To eliminate the potential impact of inter-session variations on the experimental results, we choose data from a single session for comparative validation.

### C. Signal Processing

The Root Mean Square (RMS) of SEMG is associated with factors such as the count of active muscle fibers, the pace of activation, and the isometric contraction force within a short time [72]. Compared to rectified sEMG signals, using the RMS filter can effectively minimise amplitude cancelation in the signal [73]. Therefore, a moving RMS filter of 100 milliseconds was performed during GAN model training to obtain low-frequency sEMG activation signals in this work [74]. As the relatively smooth characteristics of IMU signals (ACC signals and Euler angle signals in this work) compared to sEMG signals, and to avoid introducing excessive computational complexity, a moving average filter of 100 milliseconds was performed on raw IMU signals during GAN model training to ensure that the sample size of IMU signals aligns with that of sEMG signals [75, 76]. In addition, we further processed the absolute value of the raw sEMG signals from each electrode separately by applying a first-order low-pass Butterworth filter (with a cut-off frequency of 1 Hz) in HGR experiments. We also performed a moving average filter of 100 milliseconds on raw IMU signals to ensure a fair comparison with the virtual IMU signal in the performance evaluation experiment of the proposed approach. Due to memory limitations, we performed a downsampling process, reducing both the sEMG signals and the generated virtual IMU signals by a factor of 20 in the HGR experiments on FEMG-VPF Database, NinaPro DB2, DB3, DB7 and SIEM Database [16, 66]. To ensure real-time control, input delay was a crucial factor to be considered. A classical study suggested that the maximum allowable delay is 300 milliseconds [77]. For comparison with prior work, a slide window size of 200 milliseconds and step of 10 milliseconds was considered for segmentation of the sEMG signals and generated virtual IMU signals.

*D. Experimental Setup*

The proposed generative multimodal HGR approach was conducted using the TensorFlow 2.0 framework, implemented in Python. Data preprocessing and model training were performed on an AMAX TD21-Z2 workstation with the following hardware configuration: Ubuntu 20.04 operating system, 128 GB of RAM, Intel I9 10900X processor, and two NVIDIA Ge-Force GTX3080Ti graphics cards.

The proposed GAN-based deep generative model was trained using the Adaptive Moment Estimation (Adam) algorithm with a learning rate of 0.0002 and batch size of 64. The training epochs was set to 10000 and dropout was set to 0.2 in the discriminator. The multimodal CNN was trained using Stochastic Gradient Descent (SGD) with a batch size of 64 and epoch of 28. To accelerate convergence, a learning rate decay strategy was applied, with an initial learning rate of 0.1 and divided by 10 after the 16th and 24th epochs. To guarantee an adequate number of training samples, the training sets of all subjects were amalgamated for pretraining purposes. The specific configuration of pretraining refer to Wei et al. [16].

There were two experiments in this work, both of which are performed on the our collected FEMG-VPF database, Ninapro DB2, DB3, DB5, DB7 and SIEM database. Specifications of the two experiments listed in Table II. As shown in Table II, the data from half of the subjects was used to train the GAN model, whereas the data from the remaining subjects was used for gesture recognition in Experiment 1. Gesture recognition in Experiment 1 is conducted on only half of the subjects' data, rendering the results incomparable with previous work. To resolve this, we used the training data from all subjects to train the GAN model, then the data from all subjects was used for gesture recognition testing in Experiment 2.

TABLE II
THE TWO EXPERIMENTS IN THIS WORK

|  | Databases | GAN model | | gesture recognition | | |
| --- | --- | --- | --- | --- | --- | --- |
|  |  | Number of subjects | Trials for training | Number of subjects | Trials for training | Trials for testing |
| Experiment 1 | FEMG-VPF Database | 14 | 1, 2, 3, 4 | 14 | 1, 3 | 2, 4 |
|  | NinaPro DB2 [24] | 20 | 1, 2, 3, 4, 5, 6 | 20 | 1, 3, 4, 6 | 2, 5 |
|  | NinaPro DB3 [24] | 3 | 1, 2, 3, 4, 5, 6 | 3 | 1, 3, 4, 6 | 2, 5 |
|  | NinaPro DB5 [25] | 5 | 1, 2, 3, 4, 5, 6 | 5 | 1, 3, 4, 6 | 2, 5 |
|  | NinaPro DB7 [26] | 10 | 1, 2, 3, 4, 5, 6 | 10 | 1, 3, 4, 6 | 2, 5 |
|  | SIEM Database [66] | 10 | 1, 2, 3, 4, 5, 6 | 10 | 1, 3, 4, 6 | 2, 5 |
| Experiment 2 | FEMG-VPF Database | 28 | 1, 3 | 28 | 1, 3 | 2, 4 |
|  | NinaPro DB2 [24] | 40 | 1, 3, 4, 6 | 40 | 1, 3, 4, 6 | 2, 5 |
|  | NinaPro DB3 [24] | 6 | 1, 3, 4, 6 | 6 | 1, 3, 4, 6 | 2, 5 |
|  | NinaPro DB5 [25] | 10 | 1, 3, 4, 6 | 10 | 1, 3, 4, 6 | 2, 5 |
|  | NinaPro DB7 [26] | 20 | 1, 3, 4, 6 | 20 | 1, 3, 4, 6 | 2, 5 |
|  | SIEM Database [66] | 20 | 1, 3, 4, 6 | 20 | 1, 3, 4, 6 | 2, 5 |

*A. Evaluation of the Proposed Generative Multimodal HGR Approach*

In this subsection, we evaluate the effectiveness of our proposed generative multimodal HGR approach. We conducted a comparison between the gesture recognition performance achieved by the proposed generative multimodal HGR approach using sEMG signals and generated virtual IMU signals with the unimodal HGR approach only using sEMG signals. The sEMG-based unimodal HGR utilizes a unimodal CNN model, which is a single stream submodel of multimodal CNN model proposed in this work. Furthermore, we have also included gesture recognition results obtained by the same multimodal CNN, utilizing sEMG and real IMU signals.

Fig. 7 compared the gesture recognition accuracies of both the proposed generative multimodal HGR approach, unimodal HGR approach. The generative multimodal HGR approach achieved higher gesture recognition accuracy than did unimodal HGR approach. For Experiment 1, the proposed approach achieves accuracies of 69.67%, 87.79%, 80.13%, 94.05%, 95.06% and 72.24% on FEMG-VPF, DB2, DB3, DB5, DB7 and SIEM database, which represents an increase in accuracy of 6.74%, 10.58%, 11.43%, 2.15%, 5.25% and 3.27% compared to the unimodal HGR approach, respectively. For Experiment 2, the proposed approach achieves accuracies of 70.06%, 88.31%, 74.35%, 94.18%, 93.53% and 73.21% on FEMG-VPF, DB2, DB3, DB5, DB7 and SIEM database, which represents an increase in accuracy of 7.90%, 10.37%, 13.10%, 2.37%, 3.41% and 4.19% compared to the unimodal HGR approach, respectively.

For DB2, DB3, DB5 and DB7, the incorporation of virtual ACC signals resulted in an accuracy improvement of a maximum 13.10%. The accuracy improvement for DB5 was limited, which can be attributed to 1) the higher baseline accuracy of DB5 (exceeding 90.00%), making further improvement challenging; 2) DB5 incorporated 3-channel virtual ACC signals, whereas the other databases utilized 36-channel virtual ACC signals. For the FEMG-VPF and SIEM databases, we incorporated 3-channel virtual Euler angle signals. In addition, the FEMG-VPF database exhibited lower baseline results, which might be attributed to the quality of our database. To better simulate real-world scenarios, we did not rigorously abrade the skin and apply conductive gel during data acquisition, which might have introduced more noise [78]. Nonetheless, with incorporating the virtual Euler angle signals, a significant improvement in the results was observed (an increase of 6.74% and 7.90% respectively). Overall, the experiment results indicates that incorporating virtual IMU signals, whether ACC or Euler angles, can significantly improve the accuracy of sEMG-based HGR.

The red line above each bar in Fig. 7 represents the gesture recognition accuracy of multimodal HGR using sEMG and real IMU signals. As shown in Fig. 7, the gesture recognition performance of the proposed approach is closer to that of multimodal HGR using sEMG and real IMU signals, especially on the NinaPro DB2, DB3, DB5, and DB7 dataset.

It demonstrated that the proposed approach bridge the gap between the unimodal HGR and multimodal HGR without additional sensor hardware. Compared to virtual Euler angle signals, virtual ACC signals show performance that closely approximates the real signal. Euler angle signals remain relatively stable during gesture movements. Consequently, they may exhibit weaker intrinsic correlations with the time-varying sEMG signals, which are hardly captured by the deep generative model proposed in this work. This could potentially explain why the generated virtual Euler angle signals may not closely replicate the significant role played by real Euler angle signals in gesture recognition.

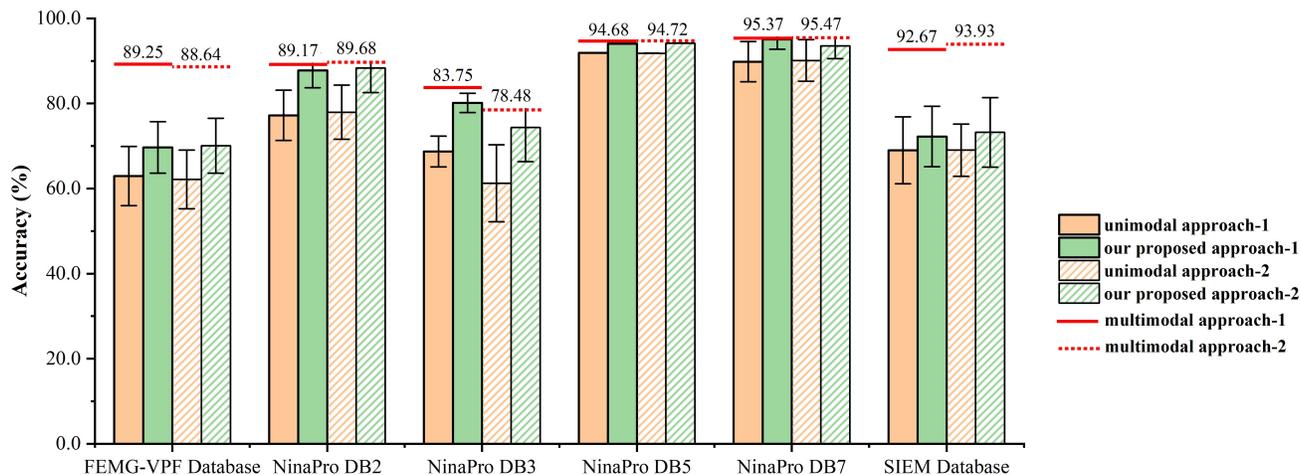

Fig. 7. Performance comparison of the our proposed generative multimodal HGR approach and unimodal HGR approach. The red line above each bar represents the gesture recognition accuracy of multimodal HGR using sEMG and real IMU signals. 1 and 2 in the legend represent Experiment 1 and Experiment 2, respectively. The height of each bar represents the average accuracy, with the error bars indicating the standard deviation.

*B. Comparison with the State-of-the-Art Gesture Recognition Approaches*

A performance comparison with state-of-the-art sEMG-based gesture recognition approaches is presented in this subsection to highlight the strengths of our proposed generative multimodal HGR approach. This comparison was conducted on four subdatabases of NinaPro. The gesture recognition accuracy of our proposed approach used here is the result on the Experiment 2. The results both of previous works and our proposed generative multimodal HGR approach are listed in Table III. To ensure consistency among subjects, the result from DB3 were solely compared to the result of MV-CNN [16], and the result from DB7 were compared to the result obtained from the 20 able-bodied subjects. As shown in Table III, the recognition accuracies of the proposed approach on various sEMG databases surpass those attained by the state-of-the-art approaches, even though some approaches focused on classifying fewer gestures or adopted longer windows. Moreover, our proposed approach directly utilizes raw sEMG signals for gesture recognition, instead of performing feature extraction on sEMG signals, as done in approaches like MV-CNN [16]. Thus, our proposed approach also offers advantages such as reduced computational load and storage requirements. The trade-off between accuracy and computation requirements is better suited and more efficient for real-time application [79].

TABLE III
GESTURE RECOGNITION ACCURACIES OF BOTH THE PROPOSED APPROACH AND EXISTING WORKS.
THE BOLD ENTRIES INDICATE THE RESULTS OF THE PROPOSED APPROACH

| Database | Approach | number of subject | number of gestures | window length ( ms) | Accuracy ( %) |
|---|---|---|---|---|---|
| NinaPro DB2 | Random Forest [24] | 40 | 50 | 200 | 75.27 |
| | ZhaiNet [80] | 40 | 50 | 200 | 78.71 |
| | HuNet [81] | 40 | 50 | 200 | 82.20 |
| | MV-CNN [16] | 40 | 50 | 200 | 83.70 |
| | TC-HGR [82] | 40 | 50 | 200 | 80.72 |
| | Our proposed approach | 40 | 50 | 200 | **88.31** |
| NinaPro DB3 | MV -CNN [16] | 6 | 50 | 200 | 64.30 |
| | Our proposed approach | 6 | 50 | 200 | **74.35** |
| NinaPro DB5 | SVM [25] | 10 | 41 | 200 | 69.04 |
| | MV-CNN [16] | 10 | 41 | 200 | 90.00 |
| | ShenNet [36] | 10 | 52 | 200 | 74.51 |
| | Alexnet [83] | 10 | 53 | 250 | 70.40 |
| | LiNet [84] | 10 | 52 | 500 | 90.82 |
| | Our proposed approach | 10 | 53 | 200 | **94.18** |
| NinaPro DB7 | LDA [26] | 20 | 40 | 256 | 60.10 |
| | MV-CNN [16] | 20 | 41 | 200 | 88.30 |
| | DNN [85] | 20 | 41 | 200 | 85.08 |
| | MMDF [86] | 20 | 40 | 256 | 91.40 |
| | Our proposed approach | 20 | 41 | 200 | **93.53** |

## VI. CONCLUSION

Multimodal gesture recognition technology is an effective means to improve the accuracy in recognizing gestures composed of hand and forearm movements, but it often results in increased hardware costs. To address this problem, we proposed a novel approach to improve sEMG-based HGR accuracy via virtual IMU signals. The virtual IMU signals was generated by a trained GAN-based deep generative model from the input sEMG signals and subsequently the sEMG signals and virtual IMU signals were fed into a multimodal CNN model for gesture recognition. Evaluations on five public databases and our collected database demonstrated that our proposed approach attained superior gesture recognition accuracies than those achieved by the sEMG-based unimodal HGR approach, with the recognition accuracy increased by 2.15%-13.10%. This indicated that incorporating virtual IMU signals can significantly improve the accuracy of sEMG-based HGR.

There are certain limitations that should be considered in future work. This study is a preliminary exploration that using generative models to bridge the gap between the unimodal HGR and multimodal HGR without additional sensor hardware. We adopted a relatively simple GAN model to generate the additional modalities. Specifically, we focused on generating ACC signals and Euler angle signals. The simplicity of the GAN model used in this work may lead to limited expressive power and less fine-grained generation of signals. Nonetheless, our proposed approach demonstrated the feasibility of improving gesture recognition accuracy through generative models. In the future, we will improve the accuracy by integrating new data modalities and using the latest generative models, such as VAE and Diffusion Model. We will also attempt to generate of finger joint angles directly, thus escaping the limitation of gesture types. This will bring a more natural and convenient experience for biomedical engineering applications such as prosthesis control, biomedical rehabilitation, robotic-assisted surgery, and human-machine interaction.

## REFERENCES

[1] L. Guo et al., "Human-Machine Interaction Sensing Technology Based on Hand Gesture Recognition: A Review," *IEEE Trans. Hum.-Mach. Syst.,* vol. 51, no. 4, pp. 300-309, Aug. 2021.
[2] B. Fang et al., "Simultaneous sEMG Recognition of Gestures and Force Levels for Interaction With Prosthetic Hand," *IEEE Trans. Neural Syst. Rehabl. Eng.,* vol. 30, pp. 2426-2436, 2022.
[3] J. Xue et al., "Underwater sEMG-based recognition of hand gestures using tensor decomposition," *Pattern Recognit. Lett,* vol. 165, pp. 39-46, Jan. 2023.
[4] X. Song et al., "Activities of Daily Living-Based Rehabilitation System for Arm and Hand Motor Function Retraining After Stroke," *IEEE Trans. Neural Syst. Rehabl. Eng.,* vol. 30, pp. 621-631, Mar. 2022.
[5] F. Xiao et al., "Human motion intention recognition method with visual, audio, and surface electromyography modalities for a mechanical hand in different environments," *Biomed. Signal Process. Control,* vol. 79, p. 104089, Jan. 2023.
[6] J. Xue and K. W. C. Lai, "Dynamic gripping force estimation and reconstruction in EMG-based human-machine interaction," *Biomed. Signal Process. Control,* vol. 80, p. 104216, Feb. 2023.
[7] M. F. Qureshi et al., "Spectral Image-Based Multiday Surface Electromyography Classification of Hand Motions Using CNN for Human–Computer Interaction," *IEEE Sensors J.,* vol. 22, no. 21, pp. 20676-20683, Nov. 2022.
[8] S. Ahmed et al., "Radar-Based Air-Writing Gesture Recognition Using a Novel Multistream CNN Approach," *IEEE Internet Things J.,* vol. 9, no. 23, pp. 23869-23880, Dec. 2022.
[9] K. Lu et al., "Channel-distribution Hybrid Deep Learning for sEMG-based Gesture Recognition," in *Proc. IEEE Int. Conf. Robot. Biomimetics*, Jinghong, China, 2022, pp. 278-284: .
[10] Y. Fang et al., "Modelling EMG driven wrist movements using a bio-inspired neural network," *Neurocomputing,* vol. 470, pp. 89-98, Jan. 2022.

[11] Y. Lin *et al.*, "Reliability Analysis for Finger Movement Recognition With Raw Electromyographic Signal by Evidential Convolutional Networks," *IEEE Trans. Neural Syst. Rehabl. Eng.*, vol. 30, pp. 96-107, Jan. 2022.
[12] D. Vera Anaya and M. R. Yuce, "Stretchable triboelectric sensor for measurement of the forearm muscles movements and fingers motion for Parkinson's disease assessment and assisting technologies," *Med. Devices Sens.*, vol. 4, no. 1, p. e10154, Feb. 2021.
[13] N. Akhlaghi *et al.*, "Real-Time Classification of Hand Motions Using Ultrasound Imaging of Forearm Muscles," *IEEE Trans. Biomed. Eng.*, vol. 63, no. 8, pp. 1687-1698, Aug. 2016.
[14] J. Yang *et al.*, "sEMG-based continuous hand gesture recognition using GMM-HMM and threshold model," in *Proc. IEEE Int. Conf. Robot. Biomimetics*, Macau, Macao, 2017, pp. 1509-1514.
[15] S. Jiang *et al.*, "Feasibility of Wrist-Worn, Real-Time Hand, and Surface Gesture Recognition via sEMG and IMU Sensing," *IEEE Trans. Ind. Informat.*, vol. 14, no. 8, pp. 3376-3385, Aug. 2018.
[16] W. Wei *et al.*, "Surface-Electromyography-Based Gesture Recognition by Multi-View Deep Learning," *IEEE Trans. Biomed. Eng.*, vol. 66, no. 10, pp. 2964-2973, Oct 2019.
[17] S. Duan *et al.*, "A Hybrid Multimodal Fusion Framework for sEMG-ACC-Based Hand Gesture Recognition," *IEEE Sensors J.*, vol. 23, no. 3, pp. 2773-2782, Feb. 2023.
[18] J. Zhang *et al.*, "Multimodal Fusion Framework Based on Statistical Attention and Contrastive Attention for Sign Language Recognition," *IEEE Trans. Mobile Comput.*, 2023.
[19] T. Y. Pan *et al.*, "A Hierarchical Hand Gesture Recognition Framework for Sports Referee Training-Based EMG and Accelerometer Sensors," *IEEE Trans. Cybern.*, vol. 52, no. 5, pp. 3172-3183, May. 2022.
[20] J. Wu *et al.*, "A Wearable System for Recognizing American Sign Language in Real-Time Using IMU and Surface EMG Sensors," *IEEE J. Biomed. Health Informat.*, vol. 20, no. 5, pp. 1281-1290, Sep. 2016.
[21] Y. Hu *et al.*, "sEMG-Based Gesture Recognition With Embedded Virtual Hand Poses and Adversarial Learning," *IEEE Access*, vol. 7, pp. 104108-104120, 2019.
[22] A. Oussidi and A. Elhassouny, "Deep generative models: Survey," in *Proc. Int. Conf. Intell. Syst. Comput. Vis.*, Fez, Morocco, 2018, pp. 1-8.
[23] I. Goodfellow *et al.*, "Generative adversarial nets," in *Proc. 27th Int. Conf. Neural Inf. Process. Syst.*, Montreal, Quebec, Canada, 2014, pp. 2672-2680.
[24] M. Atzori *et al.*, "Electromyography data for non-invasive naturally-controlled robotic hand prostheses," *Sci. Data,* vol. 1, no. 1, Dec. 2014.
[25] S. Pizzolato *et al.*, "Comparison of six electromyography acquisition setups on hand movement classification tasks," *PLoS ONE,* vol. 12, no. 10, p. e0186132, 2017.
[26] A. Krasoulis *et al.*, "Improved prosthetic hand control with concurrent use of myoelectric and inertial measurements," *J. Neuroeng. Rehabil.,* vol. 14, pp. 1-14, Jul. 2017.
[27] F. Palermo *et al.*, "Repeatability of grasp recognition for robotic hand prosthesis control based on sEMG data," *Proc. IEEE Int. Conf. Rehabil. Robot.*, pp. 1154-1159, 2017.
[28] A. Krasoulis *et al.*, "Effect of user practice on prosthetic finger control with an intuitive myoelectric decoder," *Frontiers Neurosci.*, vol. 13, p. 891, Sep. 2019.
[29] X. Jiang *et al.*, "Optimization of HD-sEMG-Based Cross-Day Hand Gesture Classification by Optimal Feature Extraction and Data Augmentation," *IEEE Trans. Hum.-Mach. Syst.*, vol. 52, no. 6, pp. 1281-1291, Dec. 2022.
[30] C. Shen *et al.*, "Toward Generalization of sEMG-Based Pattern Recognition: A Novel Feature Extraction for Gesture Recognition," *IEEE Trans. Instrum. Meas.*, vol. 71, pp. 1-12, 2022.
[31] E. Tyacke *et al.*, "Hand Gesture Recognition via Transient sEMG Using Transfer Learning of Dilated Efficient CapsNet: Towards Generalization for Neurorobotics," *IEEE Robot. Autom. Lett.*, vol. 7, no. 4, pp. 9216-9223, Oct. 2022.
[32] S. Wei *et al.*, "A Multimodal Multilevel Converged Attention Network for Hand Gesture Recognition With Hybrid sEMG and A-Mode Ultrasound Sensing," *IEEE Trans. Cybern.*, 2022.
[33] Y. Yang *et al.*, "Performance Comparison of Gesture Recognition System Based on Different Classifiers," *IEEE Trans. Cognit. Develop. Syst.*, vol. 13, no. 1, pp. 141-150, Mar. 2021.
[34] A. Fatayer *et al.*, "sEMG-Based Gesture Recognition Using Deep Learning From Noisy Labels," *IEEE J. Biomed. Health Informat.*, vol. 26, no. 9, pp. 4462-4473, Sep. 2022.
[35] Y. Zhang *et al.*, "Research on sEMG-Based Gesture Recognition by Dual-View Deep Learning," *IEEE Access*, vol. 10, pp. 32928-32937, 2022.
[36] S. Shen *et al.*, "Gesture Recognition Through sEMG with Wearable Device Based on Deep Learning," *Mobile Netw. Appl.*, vol. 25, pp. 2447–2458, Jul. 2020.
[37] W. Geng *et al.*, "Gesture recognition by instantaneous surface EMG images," *Sci. Rep.*, vol. 6, p. 36571, Nov. 2016.
[38] S. Shen *et al.*, "Movements Classification Through sEMG With Convolutional Vision Transformer and Stacking Ensemble Learning," *IEEE Sensors J.*, vol. 22, no. 13, pp. 13318-13325, Jul. 2022.
[39] H. Wang *et al.*, "sEMG based hand gesture recognition with deformable convolutional network," *Int. J. Mach. Learn. Cybern.*, vol. 13, no. 6, pp. 1729-1738, Jan. 2022.
[40] Y. Liu *et al.*, "A CNN-Transformer Hybrid Recognition Approach for sEMG-based Dynamic Gesture Prediction," *IEEE Trans. Instrum. Meas.*, vol. 72, pp. 1-16, 2023.
[41] *Ctrl-Labs: https://www.curtisbarbre.com/ctrl-kit.*
[42] *eCon: http://torintek.com/wearables/.*
[43] S. Shen *et al.*, "ICA-CNN: Gesture Recognition Using CNN With Improved Channel Attention Mechanism and Multimodal Signals," *IEEE Sensors J.*, vol. 23, no. 4, pp. 4052-4059, Feb. 2023.
[44] M. Kahng *et al.*, "GAN Lab: Understanding Complex Deep Generative Models using Interactive Visual Experimentation," *IEEE Trans. Visualization Comput. Graph.*, vol. 25, no. 1, pp. 310-320, Jan. 2019.
[45] M. Mirza and S. Osindero, "Conditional generative adversarial nets," 2014, arXiv:1411.1784.
[46] A. Radford *et al.*, "Unsupervised representation learning with deep convolutional generative adversarial networks," 2015, arXiv:1511.06434.
[47] M. Arjovsky *et al.*, "Wasserstein GAN," 2017, arXiv:1701.07875.
[48] D. P. Kingma and M. Welling, "Auto-encoding variational bayes," 2013, arXiv:1312.6114.
[49] J. Ho *et al.*, "Denoising diffusion probabilistic models," in *Proc. 34th Int. Conf. Neural Inf. Process. Syst.*, 2020, pp. 6840-6851.
[50] S. Lee *et al.*, "Progressive deblurring of diffusion models for coarse-to-fine image synthesis," 2022, arXiv:2207.11192.
[51] D. Chira *et al.*, "Image Super-Resolution with Deep Variational Autoencoders," presented at the Computer Vision – ECCV 2022 Workshops, 2023.
[52] W. Ahmad *et al.*, "A new generative adversarial network for medical images super resolution," *Sci. Rep.*, vol. 9533, Jun. 2022.
[53] S. Gong *et al.*, "Diffuseq: Sequence to sequence text generation with diffusion models," 2022, arXiv:2210.08933.
[54] Z. Kong *et al.*, "Diffwave: A versatile diffusion model for audio synthesis," 2020, arXiv:2009.09761.
[55] R. Huang *et al.*, "Make-an-audio: Text-to-audio generation with prompt-enhanced diffusion models," 2023, arXiv:2301.12661.
[56] Y. Xue *et al.*, "Human In-Hand Motion Recognition Based on Multi-Modal Perception Information Fusion," *IEEE Sensors J.*, vol. 22, no. 7, pp. 6793-6805, Apr. 2022.
[57] L. J. Ratliff *et al.*, "Characterization and computation of local Nash equilibria in continuous games," in *51st Annual Allerton Conference on Communication, Control, and Computing*, Monticello, IL, USA, 2013, pp. 917-924.
[58] K. Wang *et al.*, "Generative adversarial networks: introduction and outlook," *IEEE/CAA J. Automatica Sinica,* vol. 4, no. 4, pp. 588-598, 2017.
[59] W. Fang *et al.*, "Gesture Recognition Based on CNN and DCGAN for Calculation and Text Output," *IEEE Access,* vol. 7, pp. 28230-28237, 2019.
[60] S. Ioffe and C. Szegedy, "Batch Normalization: Accelerating Deep Network Training by Reducing Internal Covariate Shift," in *Proc. Int. Conf. Mach. Learn.*, New York, N. Y. USA, 2015, pp. 448-456.
[61] X. Glorot *et al.*, "Deep sparse rectifier neural networks," in *Proc. 14th Int. Conf. Artif. Intell. Statist.*, 2011, vol. 15, pp. 315-323.
[62] J. Xu *et al.*, "Reluplex made more practical: Leaky ReLU," in *Proc. IEEE Symp. Comput. Commun.*, Rennes, France, 2020, pp. 1-7.
[63] N. Srivastava *et al.*, "Dropout: A Simple Way to Prevent Neural Networks from Overfitting," *J. Mach. Learn. Res.*, vol. 15, pp. 1929-1958, 2014.
[64] D. J. Saunders *et al.*, "Locally connected spiking neural networks for unsupervised feature learning," *Neural Netw.*, vol. 119, pp. 332-340, Nov. 2019.
[65] H. Ci *et al.*, "Locally Connected Network for Monocular 3D Human Pose Estimation," *IEEE Trans. Pattern Anal. Mach. Intell.,* vol. 44, no.

3, pp. 1429-1442, Mar. 2022.
[66] X. Hu *et al.*, "Finger Movement Recognition via High-Density Electromyography of Intrinsic and Extrinsic Hand Muscles," *Sci. Data,* vol. 9, no. 1, p. 373, Jun. 2022.
[67] L. Ren and W. Wei, "Forearm Gesture Dataset: Gesture Recognition under Different Arm Postures and Force Levels," *figshare. Dataset,* Aug. 2023.
[68] S. Tanzarella *et al.*, "Non-invasive analysis of motor neurons controlling the intrinsic and extrinsic muscles of the hand," *J. Neural Eng.,* vol. 17, p. 046033, Aug. 2020.
[69] U. Côté-Allard *et al.*, "Transfer learning for sEMG hand gestures recognition using convolutional neural networks," in *Proc. IEEE Int. Conf. Syst. Man Cybern.*, Banff, AB, Canada, 2017, no. 1663-1668.
[70] J. L. Betthauser *et al.*, "Stable Responsive EMG Sequence Prediction and Adaptive Reinforcement With Temporal Convolutional Networks," *IEEE Trans. Biomed. Eng.,* vol. 67, no. 6, pp. 1707-1717, Jun. 2020.
[71] L. E. Osborn *et al.*, "Extended home use of an advanced osseointegrated prosthetic arm improves function, performance, and control efficiency," *J. Neural Eng.,* vol. 18, no. 2, Mar. 2021.
[72] L. Xu *et al.*, "Gesture recognition using dual-stream CNN based on fusion of sEMG energy kernel phase portrait and IMU amplitude image," *Biomed. Signal Process. Control,* vol. 73, p. 103364, Mar. 2022.
[73] G. S. Trajano *et al.*, "Neurophysiological Mechanisms Underpinning Stretch-Induced Force Loss," *Sports Med.,* vol. 47, pp. 1531–1541, Jan. 2017.
[74] Y. Yang *et al.*, "The Spiking Rates Inspired Encoder and Decoder for Spiking Neural Networks: An Illustration of Hand Gesture Recognition," *Cognitive Computation,* May. 2022.
[75] Z. Lu *et al.*, "A Hand Gesture Recognition Framework and Wearable Gesture-Based Interaction Prototype for Mobile Devices," *IEEE Trans. Hum.-Mach. Syst.,* vol. 44, no. 2, pp. 293-299, Apr. 2014.
[76] Y. L. Hsu *et al.*, "An Inertial Pen With Dynamic Time Warping Recognizer for Handwriting and Gesture Recognition," *IEEE Sensors J.,* vol. 15, no. 1, pp. 154-163, Jan. 2015.
[77] B. Hudgins *et al.*, "A new strategy for multifunction myoelectric control," *IEEE Trans. Biomed. Eng.,* vol. 40, no. 1, pp. 82-94, Jan. 1993.
[78] L. Tian *et al.*, "Large-area MRI-compatible epidermal electronic interfaces for prosthetic control and cognitive monitoring," *Nat. Biomed. Eng,,* vol. 3, pp. 194-205, Feb. 2019.
[79] C. Xu *et al.*, "Improving dynamic gesture recognition in untrimmed videos by an online lightweight framework and a new gesture dataset ZJUGesture," *Neurocomputing,* vol. 528, pp. 58-68, Feb. 2023.
[80] X. Zhai *et al.*, "Self-Recalibrating Surface EMG Pattern Recognition for Neuroprosthesis Control Based on Convolutional Neural Network," *Frontiers Neurosci.,* vol. 11, p. 379, Jul. 2017.
[81] Y. Hu *et al.*, "A novel attention-based hybrid CNN-RNN architecture for sEMG-based gesture recognition," *PLoS ONE,* vol. 13, no. 10, p. e0206049, 2018.
[82] E. Rahimian *et al.*, "Hand Gesture Recognition Using Temporal Convolutions and Attention Mechanism," in *IEEE International Conference on Acoustics, Speech and Signal Processing*, Singapore, 2022, pp. 1196-1200.
[83] Y. Li *et al.*, "Transfer Learning-Based Muscle Activity Decoding Scheme by Low-frequency sEMG for Wearable Low-cost Application," *IEEE Access,* vol. 9, pp. 22804-22815, 2021.
[84] Y. Li *et al.*, "Gesture Recognition Based on EEMD and Cosine Laplacian Eigenmap," *IEEE Sensors J.,* vol. 23, no. 14, pp. 16332-16342, Jul. 2023.
[85] P. Sri-Iesaranusorn *et al.*, "Classification of 41 Hand and Wrist Movements via Surface Electromyogram Using Deep Neural Network," *Frontiers Bioeng. Biotechnol.,* vol. 9, p. 548357, Jun. 2021.
[86] Y. Fang *et al.*, "Multi-modality deep forest for hand motion recognition via fusing sEMG and acceleration signals," *Int. J. Mach. Learn. Cybern.,* Nov. 2022.